\documentclass[letterpaper, 10 pt, conference]{ieeeconf}

\IEEEoverridecommandlockouts
\usepackage{amsmath,amssymb}
\usepackage{algpseudocode,algorithm}
\usepackage{mathtools}
\usepackage{graphicx}

\usepackage{amsthm}

\newtheorem{theorem}{Theorem}[section]

\newtheorem{lemma}[theorem]{Lemma}
\newtheorem{remark}[theorem]{Remark}

\newcommand{\X}{\mathcal{X}}
\newcommand{\U}{\mathcal{U}}

\newcommand{\Xs}{\X_{\text{safe}}}

\title{\LARGE \bf
Safe Reinforcement Learning with Nonlinear Dynamics via Model Predictive Shielding
}

\author{Osbert Bastani$^{1}$%
\thanks{$^{1}$Osbert Bastani is with the Deapartment of Computer and Information Science, University of Pennsylvania}%
}

\begin{document}

\maketitle
\thispagestyle{empty}
\pagestyle{empty}

\begin{abstract}
Reinforcement learning is a promising approach to synthesizing policies for challenging robotics tasks. A key problem is how to ensure safety of the learned policy---e.g., that a walking robot does not fall over or that an autonomous car does not run into an obstacle. We focus on the setting where the dynamics are known, and the goal is to ensure that a policy trained in simulation satisfies a given safety constraint. We propose an approach, called model predictive shielding (MPS), that switches on-the-fly between a learned policy and a backup policy to ensure safety. We prove that our approach guarantees safety, and empirically evaluate it on the cart-pole.

\end{abstract}

\section{Introduction}
\label{sec:intro}

Reinforcement learning has recently proven to be a promising approach for synthesizing neural network control policies for accomplishing challenging control tasks. We focus on the planning setting with known and deterministic dynamics---in this setting, reinforcement learning is can be used to learn policy in a simulator, and the goal is to deploy this policy to control a real robot. For instance, this approach has been used to automatically synthesize policies for challenging control problems such as object manipulation~\cite{andrychowicz2018learning} and multi-agent control~\cite{khan2019graph}, or to compress a computationally expensive search-based planner or optimal controller into a neural network policy that is computationally efficient in comparison~\cite{levine2013guided}.

A major challenge for deploying learned policies on real robots is how to guarantee that the learned policy satisfies given safety constraints. In optimal control, the generated controls are guaranteed to satisfy the safety constraints (assuming a feasible solution exists), yet reinforcement learning cannot currently provide these kinds of guarantees. Furthermore, we assume that while the environment may not be known ahead-of-time, perception is accurate, so we know the positions of the obstacles when executing the policy. As a concrete example, consider an autonomous car. We have very good models of car dynamics, and we have good sensors for detecting obstacles. However, we may want the car to drive in many different environments, with different configurations of obstacles (e.g., walls, buildings, and trees). Given a learned policy, our goal is to ensure that the policy does not cause an accident when driving in a novel environment.

One approach to guaranteeing safety is to rely on ahead-of-time verification---i.e., prove ahead-of-time that the learned policy is safe, and then deploy the learned policy on the robot~\cite{bastani2018verifiable,ivanov2019verisig}. A related approach, called \emph{shielding}, is to synthesize a backup policy and prove that it is safe, and then use the backup policy to override the learned policy as needed to guarantee safety~\cite{perkins2002lypaunov,gillula2012guaranteed,akametalu2014reachability,chow2018lyapunov,alshiekh2018safe,zhu2019inductive}. However these approaches can be computationally intractable for high-dimensional state spaces. This can be a major problem for robots operating in open world environments---in particular, to handle novel environments, we must encode the environment in the state, which can quickly increase the dimension of the state space.

An alternative approach to safe reinforcement learning is to verify safety on-the-fly. One recently proposed approach in this direction is \emph{model predictive safety certification (MPSC)}~\cite{wabersich2018linear,wabersich2019probabilistic}. This approach ensures recursive feasibility by using a model predictive controller (MPC) to ensure that the next state visited by the learned policy. If the MPC for the next state is feasible, then it uses the learned policy. Otherwise, it switches to using the MPC on the current state. Either way, the MPC is guaranteed to be feasible for the next state visited, so the overall controller is recursively feasible. However, these approaches have focused on linear dynamical systems where constrained MPC is computationally tractable.

We propose an approach that verifies safety on-the-fly that generalizes to deterministic nonlinear dynamical systems. Our approach, which we call \emph{model predictive shielding (MPS)}, is based on the concept of shielding. In contrast to ahead-of-time verification, our algorithm chooses whether to use the learned policy or the backup policy on-the-fly. At a high level, MPS maintains the invariant that the backup policy can always recover the robot, and only uses the learned policy if it can prove that doing so maintains this invariant. It checks whether this invariant holds on-the-fly by simulating the dynamics. Intuitively, checking the invariant for just the current state is far more efficient than verifying ahead-of-time that safety holds for all initial states. While our approach incurs runtime overhead during computation of the policy, each computation is efficient. In contrast, ahead-of-time verification can take exponential time; even though this computation is offline, it can be infeasible for problems of interest.

A key challenge is how to construct the backup policy. We propose an approach that decomposes it into two parts: (i) an \emph{invariant policy}, which stabilizes the robot near a safe equilibrium point (additionally, for unstable equilibrium, we propose to use feedback control to stabilize the robot), and (ii) a \emph{recovery policy}, which tries to drive the robot to a safe equilibrium point. Thus, in contrast to MPSC, in our approach, the recovery policy can be arbitrary---e.g., it can itself be trained using reinforcement learning. Thus, we can ensure it is computationally efficient even for nonlinear dynamical systems.\footnote{The invariant policy is simply a linear feedback policy, so it is also computationally efficient.}

\paragraph{Example}

Consider a walking robot, where the goal is to have the robot run as fast as possible without falling over. The learned policy may perform well at this task, but cannot guarantee safety. We consider equilibrium points where the robot is standing upright at rest. Then, the equilibrium policy stabilizes the robot at these points, and the recovery policy is trained to bring the running robot to a stop. Finally, the MPS algorithm uses the learned policy to run, while maintaining the invariant that the recovery policy can always safely bring the robot to a stop, after which the equilibrium policy can ensure safety for an infinite horizon. A key feature of our approach is that it naturally switches between the learned and backup policies. For example, suppose our algorithm uses the recovery policy to slow down the robot. The robot does not have to come to a stop; instead, our algorithm switches back to the learned policy as soon as it is safe to do so.

\paragraph{Contributions}

We propose a new algorithm for ensuring safety of a learned control policy (Section~\ref{sec:algo}), we propose an approach for constructing a backup policy in this setting (Section~\ref{sec:backup}), along with an extension to handle unstable equilibrium points (Section~\ref{sec:unstable}), and we empirically demonstrate the benefits of our approach compared to ones based on ahead-of-time verification (Section~\ref{sec:exp}).

\section{Related Work}

There has been much recent interest in safe reinforcement learning~\cite{garcia2015comprehensive,amodei2016}. One approach is to use constrained reinforcement learning to learn policies that satisfy a safety constraint~\cite{achiam2017constrained,wen2018constrained}. However, these approaches typically do not guarantee safety.

Existing approaches that guarantee safety typically rely on proving ahead-of-time that the safety property
\begin{align*}
\phi_{\text{safe}}=\bigwedge_{x_0\in\mathcal{X}_0}\bigwedge_{t=0}^{\infty}x_t\in\mathcal{X}_{\text{safe}}
\end{align*}
holds, where $\mathcal{X}_0$ are the initial states, $x_{t+1}=f(x_t,\pi(x_t))$ for all $t\ge0$, and $\mathcal{X}_{\text{safe}}$ are the safe states. One approach is to directly verify that the learned policy is safe~\cite{berkenkamp2017safe,verma2018programmatically,bastani2018verifiable,ivanov2019verisig}. However, verification does not give a way to repair the learned policy if it turns out to be unsafe.

An alternative approach, called \emph{shielding}, is use ahead-of-time verification to prove safety for a \emph{backup policy}, and then combine the learned policy with the backup policy in a way that is guaranteed to be safe~\cite{perkins2002lypaunov,gillula2012guaranteed,akametalu2014reachability,chow2018lyapunov,alshiekh2018safe,zhu2019inductive}.\footnote{More generally, the shield can simply constrain the set of allowed actions in a way that ensures safety.}
This approach can improve scalability since the backup policy is often simpler than the learned policy. For example, the backup policy may bring the to a stop if it goes near an obstacle. This approach implicitly verifies safety of the joint policy (i.e., the combination of the learned policy and the backup policy) ahead-of-time.

However, ahead-of-time verification can be computationally infeasible---it requires checking whether safety holds from every state, which can scale exponentially in the state space dimension. Many existing approaches only scale to a few dimensions~\cite{gillula2012guaranteed,berkenkamp2017safe}. One solution is to overapproximate the dynamics~\cite{asselborn2013control,koller2018learning}. However, for nonlinear dynamics, the approximation error quickly compounds, causing verification to fail even when safety holds. Scalability is particularly challenging when we want to handle the possibility of novel environments. One way to handle novel environments is to run verification from scratch every time a novel environment is encountered; however, doing so online would be computationally expensive. Our approach to handling novel environments is to encode the environment into the state. However, this approach quickly increases the dimension of the state space, resulting in poor scalability for existing approaches since they rely on ahead-of-time verification. Instead, these approaches typically focus on verifying a property of the robot dynamics in isolation of its environment (e.g., positions of obstacles) or with respect to a fixed environment. 

Finally, while we focus on planning, where the dynamics are known, there has also been work on safe exploration, which aims to ensure safety while learning the dynamics~\cite{moldovan2012safe,gillula2012guaranteed,akametalu2014reachability,turchetta2016safe,wu2016conservative,berkenkamp2017safe,dean2018safely}. These approaches rely on verification, so we believe our approach can benefit them as well.

\section{Model Predictive Shielding}
\label{sec:algo}

Given an arbitrary \emph{learned policy} $\hat{\pi}$ (designed to minimize a loss function), our goal is to minimally modify $\hat{\pi}$ to obtain a safe policy $\pi_{\text{shield}}$ for which safety is guaranteed to hold. At a high level, our algorithm ensures safety by combining $\hat{\pi}$ with a \emph{backup policy} $\pi_{\text{backup}}$ (guaranteed to ensure safety on a subset of states).

As a running example, for the cart-pole, $\hat{\pi}$ may be learned using reinforcement learning to move the cart as quickly as possible to the right, but cannot guarantee that the desired safety property that pole does not fall over. In contrast, $\pi_{\text{backup}}$ may try to stabilize the pole in place, but does not move the cart to the right.

In general, \emph{shielding} is an approach to safety based on constructing a policy $\pi_{\text{shield}}$ that chooses between using $\hat{\pi}$ and using $\pi_{\text{backup}}$. Our shielding algorithm, called \emph{model predictive shielding (MPS)}, maintains the invariant that $\pi_{\text{backup}}$ can be used to ensure safety. In particular, given a state $x$, we simulate the dynamics to determine the state $x'$ reached by using $\hat{\pi}$ at $x$, and then further simulate the dynamics to determine whether $\pi_{\text{backup}}$ can ensure safety from $x'$ using $\pi_{\text{backup}}$. For now, we describe our approach assuming $\pi_{\text{backup}}$ is given; in Sections~\ref{sec:backup} \&~\ref{sec:unstable}, we describe approaches for constructing $\pi_{\text{backup}}$.

\paragraph{Preliminaries}

We consider deterministic, discrete time dynamics $f:\mathcal{X}\times\mathcal{U}\to\mathcal{X}$ with states $\X\subseteq\mathbb{R}^{n_X}$ and actions $\U\subseteq\mathbb{R}^{n_U}$. Given a control policy $\pi:\X\to\U$, $f^{(\pi)}(x)=f(x,\pi(x))$ denotes the closed-loop dynamics. The \emph{trajectory} generated by $\pi$ from an initial state $x_0\in\mathcal{X}$ is the infinite sequence of states $x_0,x_1,...$, where $x_{t+1}=f^{(\pi)}(x_t)$ for all $t\ge0$.

\paragraph{Shielding problem}

We have two goals: (i) given loss $\ell:\mathcal{X}\times\mathcal{U}\to\mathbb{R}$, initial states $\mathcal{X}_0\subseteq\mathcal{X}$, and initial state distribution $d_0$ over $\mathcal{X}_0$, minimize
\begin{align*}
L(\pi)=\mathbb{E}_{x_0\sim d_0}\left[\sum_{t=0}^{T-1}\ell(x_t,u_t)\right],
\end{align*}
where $x_{t+1}=f(x_t,u_t)$, $u_t=\pi(x_t)$, and $T\in\mathbb{N}$ is a finite time horizon, and (ii) given safe states $\mathcal{X}_{\text{safe}}\subseteq\mathcal{X}$, ensure that the trajectory $x_0,x_1,...$ generated by $\pi$ from any $x_0\in\mathcal{X}_0$ is \emph{safe}---i.e., $x_t\in\mathcal{X}$ for all $t\ge0$.

To achieve these goals, we assume given two policies: (i) a \emph{learned policy} $\hat{\pi}$ trained to minimize $L(\pi)$, and (ii) a \emph{backup policy} $\pi_{\text{backup}}$, together with \emph{invariant states} $\mathcal{X}_{\text{inv}}\subseteq\mathcal{X}$, such that the trajectory generated by $\pi_{\text{backup}}$ from any $x_0\in\mathcal{X}_{\text{inv}}$ is guaranteed to be safe. We make no assumptions about $\hat{\pi}$; e.g., it can be a neural network policy trained using reinforcement learning. In contrast, $\pi_{\text{backup}}$ cannot be arbitrary; we give a general construction in Section~\ref{sec:backup}.

The \emph{shielding problem} is to design a policy $\pi_{\text{shield}}$ that combines $\hat{\pi}$ and $\pi_{\text{backup}}$ (i.e., $\pi_{\text{shield}}(x)\in\{\hat{\pi}(x),\pi_{\text{backup}}(x)\}$) in a way that (i) uses $\hat\pi$ as frequently as possible, and (ii) the trajectory generated by $\pi_{\text{shield}}$ from any $x_0\in\mathcal{X}_0$ is safe. Specifically, we must guarantee (ii), but not necessarily (i)---i.e., $\pi_{\text{shield}}$ must be safe, but may be suboptimal. The key challenge is deciding when to use $\hat{\pi}$ and when to use $\pi_{\text{backup}}$.

Finally, to guarantee safety, we must make some assumption about $\mathcal{X}_0$; we assume $\mathcal{X}_0\subseteq\mathcal{X}_{\text{inv}}$.

\begin{algorithm}[t]
\begin{algorithmic}
\Procedure{MPS}{$x$}
\If{$\textsc{IsRecoverable}(f^{(\hat{\pi})}(x))$}
\State \Return $\hat{\pi}(x)$
\Else
\State \Return $\pi_{\text{backup}}(x)$
\EndIf
\EndProcedure
\Procedure{IsRecoverable}{$x$}
\For{$t\in\{0,1,...,N-1\}$}
\If{$x\in\mathcal{X}_{\text{inv}}$}
\State \Return {\bf true}
\ElsIf{$x\not\in\Xs$}
\State \Return {\bf false}
\EndIf
\State $x\gets f^{(\pi_{\text{backup}})}(x)$
\EndFor
\State \Return {\bf false}
\EndProcedure
\end{algorithmic}
\caption{Model predictive shielding (MPS).}
\label{alg:shield}
\end{algorithm}

\paragraph{Model predictive shielding (MPS)}

Our algorithm for computing $\pi_{\text{shield}}(x)$ is shown in Algorithm~\ref{alg:shield}. At a high level, it checks whether $\pi_{\text{backup}}$ can ensure safety from the state $x'=f^{(\pi)}(x)$ that would be reached by $\hat{\pi}$. If so, then it uses $\hat{\pi}$; otherwise, it uses $\pi_{\text{backup}}$.

More precisely, let $N\in\mathbb{N}$ be given. Then, a state $x\in\mathcal{X}$ is \emph{recoverable} if for the trajectory $x_0,x_1,...$ generated by $\pi_{\text{backup}}$ from $x_0=x$, there exists $t\in\{0,1,...,N-1\}$ such that (i) $x_i\in\mathcal{X}_{\text{safe}}$ for all $i\le t$, and (ii) $x_t\in\mathcal{X}_{\text{inv}}$. Intuitively, $\pi_{\text{backup}}$ safely drives the robot into an invariant state from $x$ within $N$ steps. In Algorithm~\ref{alg:shield}, \textsc{IsRecoverable} checks whether $x\in\mathcal{X}_{\text{rec}}$.

Then, $\pi_{\text{shield}}$ uses $\hat{\pi}$ if $f^{(\hat{\pi})}(x)$ is recoverable; otherwise, it uses $\pi_{\text{backup}}$. We have the following:
\begin{theorem}
\label{thm:mps}
The trajectory generated by $\pi_{\text{shield}}$ from any $x_0\in\mathcal{X}_0$ is safe.
\end{theorem}
We give proofs in Appendix~\ref{sec:proofs}.

\begin{remark}
\rm
The running time of our algorithm on each step is $O(N)$ due to the call to \textsc{IsRecoverable} (assuming $\hat{\pi}$ and $\pi_{\text{backup}}$ run in constant time). We believe this overhead is reasonable in many settings; if necessary, we can safely add a time out, and have \textsc{IsRecoverable} return false if it runs out of time.
\end{remark}

\section{Backup Policies}
\label{sec:backup}

We now discuss how to construct $\pi_{\text{backup}}$. Our construction relies on \emph{safe equilibrium points} of $f$---i.e., where the robot remains safely at rest. Most robots of interest have such equilibria---for example, the cart-pole has equilibrium points when the cart and pole are motionless, and the pole is perfectly upright. Other examples of equilibrium points include a walking robot standing upright, a quadcopter hovering at a position, or a swimming robot treading water.

One challenge is that these equilibria may be unstable; while the approach described in this section technically ensures safety, it is very sensitive to even tiny perturbations. For example, in the case of cart-pole, a tiny perturbation would cause the pole to fall down. We describe how a way to address this issue in Section~\ref{sec:unstable}.

At a high level, our backup policy $\pi_{\text{backup}}$ is composed of two policies: (i) an \emph{equilibrium policy} $\pi_{\text{eq}}$ that ensures safety at equilibrium points, and (ii) a \emph{recovery policy} $\pi_{\text{rec}}$ that tries to drive the robot to a safe equilibrium point. Then, $\pi_{\text{backup}}$ uses $\pi_{\text{rec}}$ until it reaches a safe equilibrium point, after which it uses $\pi_{\text{eq}}$. Continuing our example, for cart-pole, $\pi_{\text{rec}}$ would try to get the pole into an upright position, and then $\pi_{\text{eq}}$ would stabilize the robot near that position. We begin by describing how we construct $\pi_{\text{eq}}$ and $\pi_{\text{rec}}$, and then describe how they are combined to form $\pi_{\text{backup}}$.

\subsection{Equilibrium Policy}

An \emph{safe equilibrium point} $z\in\mathcal{Z}_{\text{eq}}\subseteq\mathcal{X}\times\mathcal{U}$ is a pair $z=(x,u)$ such that (i) $x=f(x,u)$, and (ii) $x\in\mathcal{X}_{\text{safe}}$. We let
\begin{align*}
\mathcal{X}_{\text{inv}}=\{x\in\mathcal{X}\mid\exists u\in\mathcal{U}~\text{s.t.}~(x,u)\in\mathcal{Z}_{\text{eq}}\}.
\end{align*}
Furthermore, for $(x,u)\in\mathcal{Z}_{\text{eq}}$, we let $\pi_{\text{eq}}(x)=u$; if multiple such $u$ exist, we pick an arbitrary one. Then, $\pi_{\text{eq}}$ and $\mathcal{X}_{\text{inv}}$ satisfy the conditions for the backup policy. As we describe below, we do not need to define $\pi_{\text{eq}}$ outside of $\mathcal{X}_{\text{inv}}$.

\subsection{Recovery Policy}

Using $\pi_{\text{backup}}=\pi_{\text{eq}}$ can result in poor performance. In particular, $\pi_{\text{backup}}$ it is undefined outside of $\mathcal{X}_{\text{inv}}$, so $\mathcal{X}_{\text{rec}}=\mathcal{X}_{\text{inv}}$. As a consequence, $\pi_{\text{shield}}$ will keep the robot inside $\mathcal{X}_{\text{inv}}$. However, since $\mathcal{X}_{\text{inv}}$ consists of equilibrium points, the robot will never move.

Thus, we additionally train a \emph{recovery policy} $\pi_{\text{rec}}$ that attempts to drive the robot into $\mathcal{X}_{\text{inv}}$. The choice of $\pi_{\text{rec}}$ can be arbitrary; however, $\pi_{\text{shield}}$ achieves lower loss for better $\pi_{\text{rec}}$. There is sometimes an obvious choice (e.g., for a autonomous car, $\pi_{\text{rec}}$ may simply slam the brakes), but not always.

In general, we can use reinforcement learning to train $\pi_{\text{rec}}$. At a high level, we train it to drive the robot from a safe state reached by $\hat{\pi}$ to the closest safe equilibrium point. First, we use initial state distribution $d_{\text{rec}}$; we define $d_{\text{rec}}$ by describing how to take a single sample $x\sim d_{\text{rec}}$: (i) sample an initial state $x_0\sim d_0$, (ii) sample a time horizon $t\sim\text{Uniform}(\{0,...,N\})$, (iii) compute the trajectory $x_0,x_1,...$ generated by $\hat{\pi}$ from $x_0$, and (iv) reject if $x_t\not\in\mathcal{X}_{\text{safe}}$; otherwise take $x=x_t$. Second, we use loss $\ell_{\text{rec}}(x,u)=-\mathbb{I}[x\in\mathcal{X}_{\text{inv}}]$, where $\mathbb{I}$ is the indicator function. We can also use a shaped loss---e.g., $\ell_{\text{rec}}(x,u)=\|x-x'\|^2$, where $x'\in\mathcal{X}_{\text{inv}}$ is the closest safe equilibrium point. Then, we use reinforcement learning to train
\begin{align*}
\pi_{\text{rec}}=\operatorname*{\arg\min}_{\pi}\mathbb{E}_{x_0\sim d_{\text{rec}}}\left[\sum_{t=0}^{T-1}\ell_{\text{rec}}(x_t,u_t)\right],
\end{align*}
where $x_{t+1}=f(x_t,u_t)$, $u_t=\pi(x_t)$, and $T\in\mathbb{N}$.

\subsection{Backup Policy}

Finally, we have
\begin{align*}
\pi_{\text{backup}}(x)=\begin{cases}\pi_{\text{eq}}(x)&\text{if}~x\in\mathcal{X}_{\text{inv}}\\\pi_{\text{rec}}(x)&\text{otherwise}.\end{cases}
\end{align*}
By construction, $\pi_{\text{backup}}$ and $\mathcal{X}_{\text{inv}}$ satisfy the conditions for a backup policy.

\section{Unstable Equilibrium Points}
\label{sec:unstable}

For unstable equilibria $z\in\mathcal{Z}_{\text{eq}}$, we use feedback stabilization to ensure safety. As in Section~\ref{sec:backup}, $\pi_{\text{backup}}$ is composed of an equilibrium policy $\pi_{\text{eq}}$, which is safe on $\mathcal{X}_{\text{inv}}$, and a recovery policy $\pi_{\text{rec}}$, which tries to drive the robot to $\mathcal{X}_{\text{inv}}$. In this section, we focus on constructing $\pi_{\text{eq}}$ and $\mathcal{X}_{\text{inv}}$; we can train $\pi_{\text{rec}}$ as in Section~\ref{sec:backup}. At a high level, we choose $\pi_{\text{eq}}$ to be the LQR for the linear approximation $\tilde{f}$ of the dynamics around $z$, and then use LQR verification to compute the states $\mathcal{X}_{\text{inv}}$ for which $\pi_{\text{eq}}$ is guaranteed to be safe. We begin by giving background on LQR control and verification, and then describe our construction.

\subsection{Assumptions}

For tractability, our algorithm makes two additional assumptions. First, we assume that the dynamics $f$ is a degree $d$ polynomial.
\footnote{In particular, $f$ is a multivariate polynomial over $x\in\X$ with real coefficients.}
Second, we assume that the safe set is a convex polytope---i.e.,
\begin{align*}
\mathcal{X}_{\text{safe}}=\{x\in\mathcal{X}\mid A_{\text{safe}}x\le b_{\text{safe}}\},
\end{align*}
where $A_{\text{safe}}\in\mathbb{R}^{k\times n_X}$ and $b_{\text{safe}}\in\mathbb{R}^k$ for some $k\in\mathbb{N}$.

\begin{remark}
\rm
As in prior work~\cite{tedrake2009lqr}, for non-polynomial dynamics, we use local Taylor approximations; while our theoretical safety guarantees do not hold, safety holds in practice since these approximations are very accurate. Furthermore, as described below, our results easily extend to arbitrary $\mathcal{X}_{\text{safe}}$.
\end{remark}

\subsection{LQR control}

Consider linear dynamics $\tilde{f}(x,u)=Ax+Bu$, where $A\in\mathbb{R}^{n_X\times n_X}$ and $B\in\mathbb{R}^{n_X\times n_U}$, with loss $\ell(x,u)=x^{\top}Qx+u^{\top}Ru$, where $Q\in\mathbb{R}^{n_X\times n_X}$ and $R\in\mathbb{R}^{n_U\times n_U}$. Then, the optimal policy for these dynamics is a linear policy $\pi_{\text{eq}}(x)=Kx$, where $K\in\mathbb{R}^{n_U\times n_X}$, called the linear quadratic regulator (LQR)~\cite{tedrake2018underactuated}.
\footnote{The LQR is optimal for the infinite horizon problem.}
Additionally, the cost-to-go function (i.e., the negative value function) of the LQR has the form $J(x)=x^{\top}Px$, where $P\in\mathbb{R}^{n_X\times n_X}$ is a positive semidefinite matrix. Both the LQR and its cost-to-go can be computed efficiently~\cite{tedrake2018underactuated}.

To stabilize the robot near $z\in\mathcal{Z}_{\text{eq}}$, we use the LQR $\pi_{\text{eq}}$ for the linear approximation $\tilde{f}$ of $f$ around $z$; the cost matrices $Q,R$ can each be chosen to be any positive definite matrix---e.g., the identity. Since $\tilde{f}$ becomes arbitrarily accurate close to $z$, we intuitively expect $\pi_{\text{eq}}$ to be a good control policy.

\subsection{LQR Verification}

We can use LQR verification to compute a region around $(x,u)$ where $\pi_{\text{eq}}$ is guaranteed to be safe for an infinite horizon~\cite{parrilo2000structured,tedrake2009lqr,tedrake2018underactuated}. Given a policy $\pi$, $\mathcal{G}\subseteq\X$ is \emph{invariant} for $\pi$ if for any initial state $x_0\in\mathcal{G}$, the trajectory generated by $\pi$ from $x_0$ is contained in $\mathcal{G}$---i.e., if the robot starts from any $x_0\in\mathcal{G}$, then it remains in $\mathcal{G}$. We have~\cite{tedrake2018underactuated}:
\begin{lemma}
\label{lem:lyapunov}
Let $\pi$ be a policy. Suppose that there exists $V:\X\to\mathbb{R}$ and $\epsilon\in\mathbb{R}$ satisfying
\begin{align*}
V(x)\ge V(f^{(\pi)}(x))\hspace{0.1in}(\forall x\in\mathcal{G}_{\epsilon}=\{x\in\X\mid V(x)\le\epsilon\}).
\end{align*}
Then, $\mathcal{G}_{\epsilon}$ is an invariant set for $\pi$.
\end{lemma}
Here, $V$ is a closely related to a Lyapunov function, though we do not need the usual constraint that $V(0)=0$ and $V(x)>0$ otherwise; this definition suffices to guarantee safety, but not Lyapunov stability. By Lemma~\ref{lem:lyapunov}, given a candidate function $V$, we can use optimization to compute $\epsilon$ such that $\mathcal{G}_{\epsilon}$ is invariant. In particular, given a set $\mathcal{F}$ of functions $f:\X\to\mathbb{R}$, a policy $\pi$, and a candidate $V$, let
\begin{align}
\label{eqn:lqrverification}
\epsilon^*=&\max_{\lambda\in\mathcal{F},\vec{\mu}\in\mathcal{F}^k,\epsilon'\in\mathbb{R}}~\epsilon \hspace{0.1in} \text{subj. to} \\
&\begin{array}{rl}
V(x)-V(f^{(\pi)}(x))+\lambda(x)(V(x)-\epsilon') \hspace{-7pt} & \ge0 \\
b_{\text{safe}}-A_{\text{safe}}x + \vec{\mu}(x)(V(x)-\epsilon') \hspace{-7pt} & \ge 0 \\
\lambda(x),~\vec{\mu}(x),~\epsilon' \hspace{-7pt} & \ge0
\end{array} \nonumber
\end{align}
where the constraints are required to hold for all $x\in\mathbb{R}^{n_X}$. We have the following~\cite{tedrake2018underactuated}:
\begin{lemma}
\label{lem:sos}
We have (i) $\mathcal{G}_{\epsilon^*}$ is invariant for $\pi$, and (ii) $\pi$ is safe from any $x_0\in\mathcal{G}_{\epsilon^*}$.
\end{lemma}

As with candidate Lyapunov functions, we can choose our candidate $V$ to be the cost-to-go function $J$ of $\pi_{\text{eq}}$---i.e., $V(x)=J(x)=x^{\top}Px$. Indeed, for the linear approximation $\tilde{f}$, $J$ is a Lyapunov function of $\pi_{\text{eq}}$ on all of $\mathbb{R}^{n_X}$. Thus, $J$ is a promising choice of the candidate for $V$ for the true dynamics $f$.

The optimization problem (\ref{eqn:lqrverification}) is intractable in general. We use a standard modification that strengthens the constraints to obtain tractability; the resulting solution is guaranteed to satisfy the original constraints, but may achieve a suboptimal objective value. First, for some $d'\in\mathbb{N}$, we choose $\mathcal{F}$ to be the set of polynomials in $x$ of degree at most $d'$. Then, for $\pi=\pi_{\text{eq}}$, each constraint in (\ref{eqn:lqrverification}) has form $p(x)\ge0$ for some polynomial $p(x)$. We replace each constraint $p(x)\ge0$ with the stronger constraint that $p(x)$ is a \emph{sum-of-squares (SOS)})---i.e., $p(x)=p_1(x)^2+...+p_k(x)^2$ for some polynomials $p_1,...,p_k$. If $p(x)$ is SOS, then $p(x)\ge0$ for all $x\in\X$. With this modification, the optimization problem (\ref{eqn:lqrverification}) is an \emph{SOS program}; for our choice of $\mathcal{F}$, it can be solved efficiently using semidefinite programming~\cite{parrilo2000structured,tedrake2009lqr,tedrake2018underactuated}.

\begin{remark}
\rm
Our approach is sound---i.e., the solution to our SOS program is guaranteed to satisfy the constraints in (\ref{eqn:lqrverification}), so the statement of Lemma~\ref{lem:sos} holds; however, our solution may be suboptimal.
\end{remark}

\begin{remark}
\rm
For general $\mathcal{X}_{\text{safe}}$, given an equilibrium point $(x,u)\in\mathcal{Z}_{\text{eq}}$, consider a convex polytope $\tilde{\mathcal{X}}_{\text{safe}}=\{x\in\mathcal{X}\mid\tilde{A}_{\text{safe}}x\le\tilde{b}_{\text{safe}}\}$ satisfying (i) $\tilde{\mathcal{X}}_{\text{safe}}\subseteq\mathcal{X}_{\text{safe}}$, and (ii) $x\in\tilde{\mathcal{X}}_{\text{safe}}$. Then, we can conservatively use $\tilde{A}_{\text{safe}},\tilde{b}_{\text{safe}}$ in place of $A_{\text{safe}},b_{\text{safe}}$ when solving the optimization problem (\ref{eqn:lqrverification}).
\end{remark}

\subsection{Equilibrium Policy}

Given a safe equilibrium point $z\in\mathcal{Z}_{\text{eq}}$, let $\pi_{\text{eq}}$ be the LQR for the linear approximation $\tilde{f}$ around $z$; then, we let $\pi_z=\pi_{\text{eq}}$. Furthermore, let $\epsilon^*$ be the solution to the SOS variant of the optimization problem (\ref{eqn:lqrverification}); then, we let $\mathcal{G}_z=\mathcal{G}_{\epsilon^*}$ be an invariant set of $\pi_z$. Now, we choose
\begin{align*}
\pi_{\text{eq}}(x)=\pi_{\rho(x)}(x)
\hspace{0.1in}\text{and}\hspace{0.1in}
\mathcal{X}_{\text{inv}}&=\bigcup_{z\in\mathcal{Z}_{\text{eq}}}\mathcal{G}_z,
\end{align*}
where $\rho(x)$ is the closest equilibrium point to $x$---i.e., $\rho(x)=\operatorname*{\arg\min}_{(x',u')\in\mathcal{Z}_{\text{eq}}}\|x-x'\|$. In other words, $\pi_{\text{eq}}(x)$ uses the LQR for the equilibrium point closest to $x$, and $\mathcal{X}_{\text{inv}}$ is the set of states in the invariant set of some equilibrium point. We have the following:
\begin{theorem}
\label{thm:backup}
The trajectory generated using $\pi_{\text{eq}}$ from any $x_0\in\mathcal{X}_{\text{inv}}$ is safe.
\end{theorem}

\begin{remark}
\rm
Computing $\pi_{\text{eq}}$ is polynomial time, but may still be costly---given $x$, we need to compute the nearest equilibrium point $z$, and then compute $\pi_z$ and $\mathcal{G}_z$. In practice, we can often precompute these. For example, for cart-pole, the dynamics are equivariant under translation. Thus, we can compute the $\pi_{z_0}(x)=K_0x$ and $\mathcal{G}_{z_0}$ for the origin $z_0=(\vec{0},\vec{0})$, and perform a change of coordinates to use these for other $z$. In particular, for any $z=(x',\vec{0})\in\mathcal{Z}_{\text{eq}}$, we have $\pi_z(x)=K_0(x-x')$ and $\mathcal{G}_z=\{x'+x\mid x\in\mathcal{G}_{z_0}\}$.
\end{remark}

\begin{figure*}
\begin{center}
\includegraphics[width=0.45\textwidth]{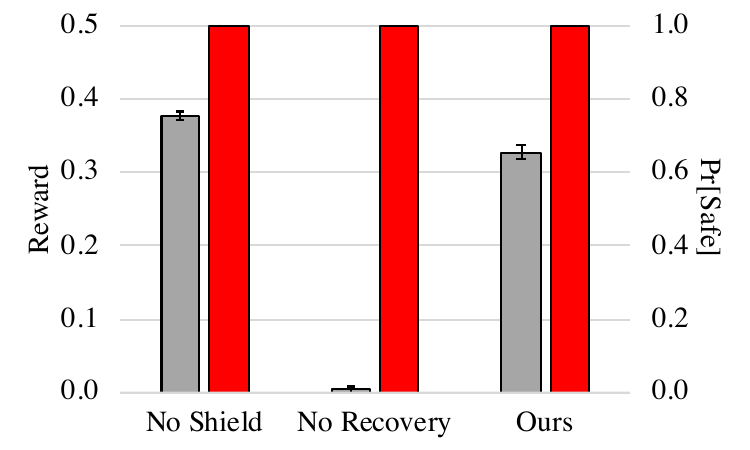}
\includegraphics[width=0.45\textwidth]{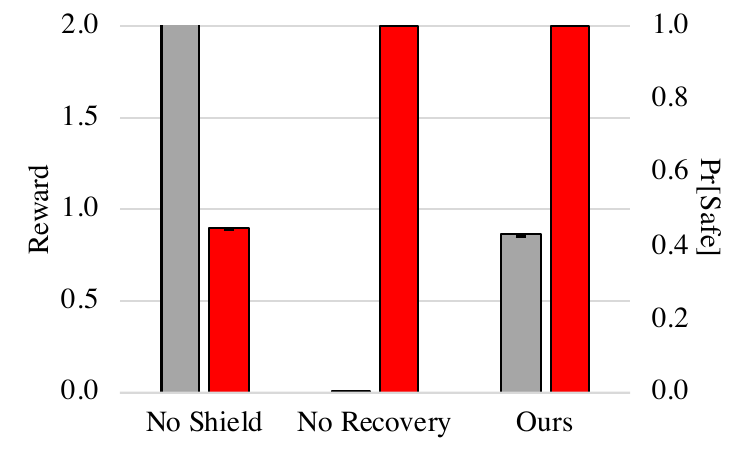}
\end{center}
\caption{Reward (gray) and safety probability (red) for original (left) and modified (middle) environments. We show means and standard errors estimated using 100 random rollouts.}
\label{fig:expperf}
\end{figure*}

\section{Experiments}
\label{sec:exp}

\subsection{Experimental Setup}

\paragraph{Benchmark}

We evaluate our approach based on the cart-pole~\cite{cartpole_problem}. In this task, the goal is to balance an inverted pole on top of a cart, where we are only able to move the cart left and right. The states are $(x,v,\theta,\omega)\in\mathbb{R}^4$ include the position and velocity of the cart, and the angle and angular velocity of the pole. The action $a\in\mathbb{R}$ is the acceleration applied to the cart. The goal is to move the cart to the right at a target velocity of $v_0=0.1$, under a safety constraint that the pole angle $\theta$ is bounded by $\theta_{\text{max}}=0.15~\text{rad}$.

\paragraph{Reinforcement learning}

We learn both $\hat{\pi}$ and $\pi_{\text{rec}}$ using backpropagation-through-time (BPTT)~\cite{bastani2020sample}. This algorithm is a model-based reinforcement learning algorithm that learns control policies by using gradient descent on the reward gradients through the dynamics. Each policy $\hat{\pi}$ and $\pi_{\text{rec}}$ is a neural network with a single hidden layer containing 200 hidden units and using ReLU activiations. For training, we randomly sample trajectories using initial states drawn uniformly at random from $[-0.05,0.05]^4$, consider a time horizon of $T=200$, and use a discount factor $\gamma=0.99$.

\paragraph{Shielding}

We use $\rho((x,v,\theta,\omega))=((x,0,0,0),0)$---i.e., stabilize the pole to the origin at the current cart position. We use the degree 5 Taylor approximation around the origin for LQR verification, and degree 6 polynomials for $\mathcal{F}$ in the SOS program. For our shield policy $\pi_{\text{shield}}$, we use a recovery horizon of $N=100$.

\paragraph{Baselines}

We compare our shield policy to two baselines. The first is using the learned policy $\hat{\pi}$ without a shield. The second, is an ablation of our shield policy $\pi_{\text{shield}}$ where the backup policy $\pi_{\text{backup}}$ includes the invariant policy $\pi_{\text{eq}}$ but not use the recovery policy $\pi_{\text{rec}}$; equivalently, it is our shield policy with a recovery horizon of $N=0$.

\paragraph{Metrics}

We consider both the reward and safety probability. First, the reward is the total distance $z$ traveled by the cart; higher is better. Second, the safety probability is the probability that a uniformly random state visited during a randomly sampled rollout is safe (i.e., $x\in\X_{\text{safe}}$).

\subsection{Results}

In Figure~\ref{fig:expperf} (left), we show the reward and safety probability achieved by our shield policy along with our two baselines. Both our shielded policy and its ablation achieve 100\% safety probability. In this case, the learned policy $\hat{\pi}$ achieves perfect safety as well, though it is not guaranteed to do so. Our shielded policy achieves lower reward than $\hat{\pi}$, but is guaranteed to be safe. Our ablation achieves the lowest reward; it is unable to move very far since it cannot leave $\mathcal{X}_{\text{inv}}$. Thus, the recovery policy is critical for achieving good performance.

\subsection{Changes in the environment}

Changes in the environment are an important cause of safety failures of learned policies. In particular, the learned policy $\hat{\pi}$ is tailored to perform well on a specific state distribution; thus, if the problem changes (for instance, different obstacle configurations or a longer time horizon), then $\hat{\pi}$ may no longer be safe to use. To demonstrate how MPS can ensure safety in the face of such changes, we consider a modification to the cart-pole where we increase the time horizon. In particular, we only $\hat{\pi}$ and $\pi_{\text{rec}}$ for rollouts of length $T=200$, but we then use them to control the cart-pole for rollouts of length $T=1000$.

In Figure~\ref{fig:expperf} (middle), we show the reward and safety probability achieved by our shield policy and our two baselines on this modified environment. As can be seen, both our shield policy and its ablation continue to achieve 100\% safety probability. In this case, the learned policy $\hat{\pi}$ achieves good performance, but it does so by achieving a very low safety probability (essentially zero).

\section{Conclusion}

We have proposed an algorithm for ensuring safety of a learned controller by composing it with a safe backup controller. Our experiments demonstrate how our approach can ensure safety without significantly sacrificing performance. We leave much room for future work---e.g., extending our approach to handle unknown dynamics, partial observability, and multi-agent robots.

\addtolength{\textheight}{-12cm}

\section*{Appendix}
\label{sec:proofs}

\textbf{Proof of Theorem~\ref{thm:mps}.}
We prove by induction that if $x_t\in\mathcal{X}_{\text{rec}}$, then $x_{t+1}=f^{(\pi_{\text{shield}})}(x_t)\in\mathcal{X}_{\text{rec}}$. The base case holds since $x_0\in\mathcal{X}_0\subseteq\mathcal{X}_{\text{stable}}\subseteq\mathcal{X}_{\text{rec}}$. For the inductive case, there are two possibilities. (i) If $x'=f^{(\hat{\pi})}(x_t)\in\mathcal{X}_{\text{rec}}$, then $\pi_{\text{shield}}(x_t)=\hat{\pi}(x_t)$, so $x_{t+1}=x'\in\mathcal{X}_{\text{rec}}$. (ii) Otherwise, $\pi_{\text{shield}}(x_t)=\pi_{\text{backup}}(x_t)$; clearly, $x_t\in\mathcal{X}_{\text{rec}}$ implies that $x''=f^{(\pi_{\text{backup}})}(x_t)\in\mathcal{X}_{\text{rec}}$, so $x_{t+1}=x''\in\mathcal{X}_{\text{rec}}$. Thus, the inductive case holds. The claim follows. $\qed$

\textbf{Proof of Lemma~\ref{lem:lyapunov}.}
The claim follows by induction. $\qed$

\textbf{Proof of Lemma~\ref{lem:sos}.}
Consider any $x$ such that $V(x)\le\epsilon$. To see (i), note that in the first constraint in (\ref{eqn:lqrverification}), the second term is negative since $\lambda(x)\ge0$, so $V(x)-V(f^{(\pi)}(x))\ge0$. Thus, by Lemma~\ref{lem:lyapunov}, $\mathcal{G}_{\epsilon}$ is invariant. Similarly, to see (ii), note that in the second constraint in (\ref{eqn:lqrverification}), the second term is negative since $\vec{\mu}(x)\ge0$, so $b_{\text{safe}}-A_{\text{safe}}x\ge0$. Thus, $x\in\Xs$ for all $x\in\mathcal{G}_{\epsilon}$. Since $\mathcal{G}_{\epsilon}$ is invariant, $\pi$ is safe from any $x_0\in\mathcal{G}_{\epsilon}$. $\qed$

\textbf{Proof of Theorem~\ref{thm:backup}.}
We prove by induction on $t$ that $x_t\in\mathcal{X}_{\text{stable}}$ for all $t\ge0$. The base case follows by assumption. For the inductive case, note that $x_t\in\mathcal{G}_{\rho(x)}$, so we have $f^{(\pi_{\rho(x)})}(x_t)\in\mathcal{G}_{\rho(x)}$ since $\mathcal{G}_{\rho(x)}$ is invariant. Thus, $x_{t+1}=f^{(\pi_{\text{backup}})}(x_t)=f^{(\pi_{\rho(x)})}(x_t)\in\mathcal{G}_{\rho(x)}$, so the inductive case follows. By construction, $\mathcal{X}_{\text{stable}}\subseteq\mathcal{X}_{\text{safe}}$. $\qed$

\bibliographystyle{IEEEtran}
\bibliography{paper}

\end{document}